\documentclass{article}
\usepackage{ijcai21}

\usepackage{times}
\usepackage{soul}
\usepackage{url}
\usepackage[hidelinks]{hyperref}
\usepackage[utf8]{inputenc}
\usepackage[small]{caption}
\usepackage{graphicx}
\usepackage{amsmath}
\usepackage{amsthm}
\usepackage{booktabs}
\usepackage{algorithm}
\usepackage{algorithmic}
\usepackage{makecell}
\usepackage{bm}
\usepackage{amsfonts,amssymb}
\usepackage{subfigure}
\usepackage[table]{xcolor}
\usepackage{color}
\usepackage{bm}
\usepackage{makecell}
\usepackage{multirow}

\title{Fairness-Aware Unsupervised Feature Selection}

\author{
Xiaoying Xing$^1$
\and
Hongfu Liu$^2$\and
Chen Chen$^{3}$\And
Jundong Li$^{3}$
\affiliations
$^1$Tsinghua University, Beijing, China 100084\\
$^2$Brandeis University, Waltham, MA, USA 02453\\
$^3$University of Virginia, Charlottesville, VA, USA 22904\\
\emails
xingxy0505@gmail.com, hongfuliu@brandeis.edu, \{zrh6du, jl6qk\}@virginia.edu
}

\begin{document}

\maketitle

\begin{abstract}
Feature selection is a prevalent data preprocessing paradigm for various learning tasks. Due to the expensive cost of acquiring supervision information, unsupervised feature selection sparks great interests recently. However, existing unsupervised feature selection algorithms do not have fairness considerations and suffer from a high risk of amplifying discrimination by selecting features that are over associated with protected attributes such as gender, race, and ethnicity. In this paper, we make an initial investigation of the fairness-aware unsupervised feature selection problem and develop a principled framework, which leverages kernel alignment to find a subset of high-quality features that can best preserve the information in the original feature space while being minimally correlated with protected attributes. Specifically, different from the mainstream in-processing debiasing methods, our proposed framework can be regarded as a model-agnostic debiasing strategy that eliminates biases and discrimination before downstream learning algorithms are involved. Experimental results on multiple real-world datasets demonstrate that our framework achieves a good trade-off between utility maximization and fairness promotion. 


\end{abstract}

\section{Introduction}
Feature selection is an effective data preprocessing strategy for a myriad of data mining and machine learning tasks~\cite{guyon2003introduction,li2017feature}. It aims to select a subset of relevant features from the original feature space while eliminating the adverse impact of irrelevant, redundant, and noisy features. In contrast to the prevalent deep learning models for representation learning, feature selection gives learning models better readability and interpretability by maintaining the physical meanings of original features, thus is often preferred in high-stake applications (e.g., clinical diagnosis~\cite{inbarani2014supervised}, employment~\cite{sobnath2020feature}, and financial analytics~\cite{liang2015effect}). According to the label availability, traditional feature selection algorithms can be mainly categorized as supervised methods and unsupervised methods~\cite{li2017feature}. As supervision information is often costly to amass in many real-world scenarios, unsupervised feature selection methods has attracted an increasing amount of attention in recent years.

Despite the successful adoption of feature selection algorithms in various high-stake decision-making scenarios, the existing selection algorithms often do not have the fairness considerations and may suffer from a risk of amplifying stereotypes and exhibiting discriminatory actions toward specific groups or populations by over associating protected attributes\footnote{We use protected and sensitive features interchangeably. Meanwhile, we also use attributes and features interchangeably.} (e.g., gender, race, and ethnicity)~\cite{chouldechova2018frontiers,du2020fairness,mehrabi2019survey,grgic2018beyond}. Although it is intuitive to manually remove these protected attributes in the selected feature subset to avoid direct discrimination, a number of non-protected attributes that are highly correlated with the protected attributes may still be selected by the algorithms and result in unintentional discrimination problems (e.g., residential zip code of a person may indicate the race information because of the population of residential areas)~\cite{zhang2016causal,kallus2019assessing}. 

\begin{figure}[!t]
\centering
\resizebox{\linewidth}{!}{
\includegraphics[]{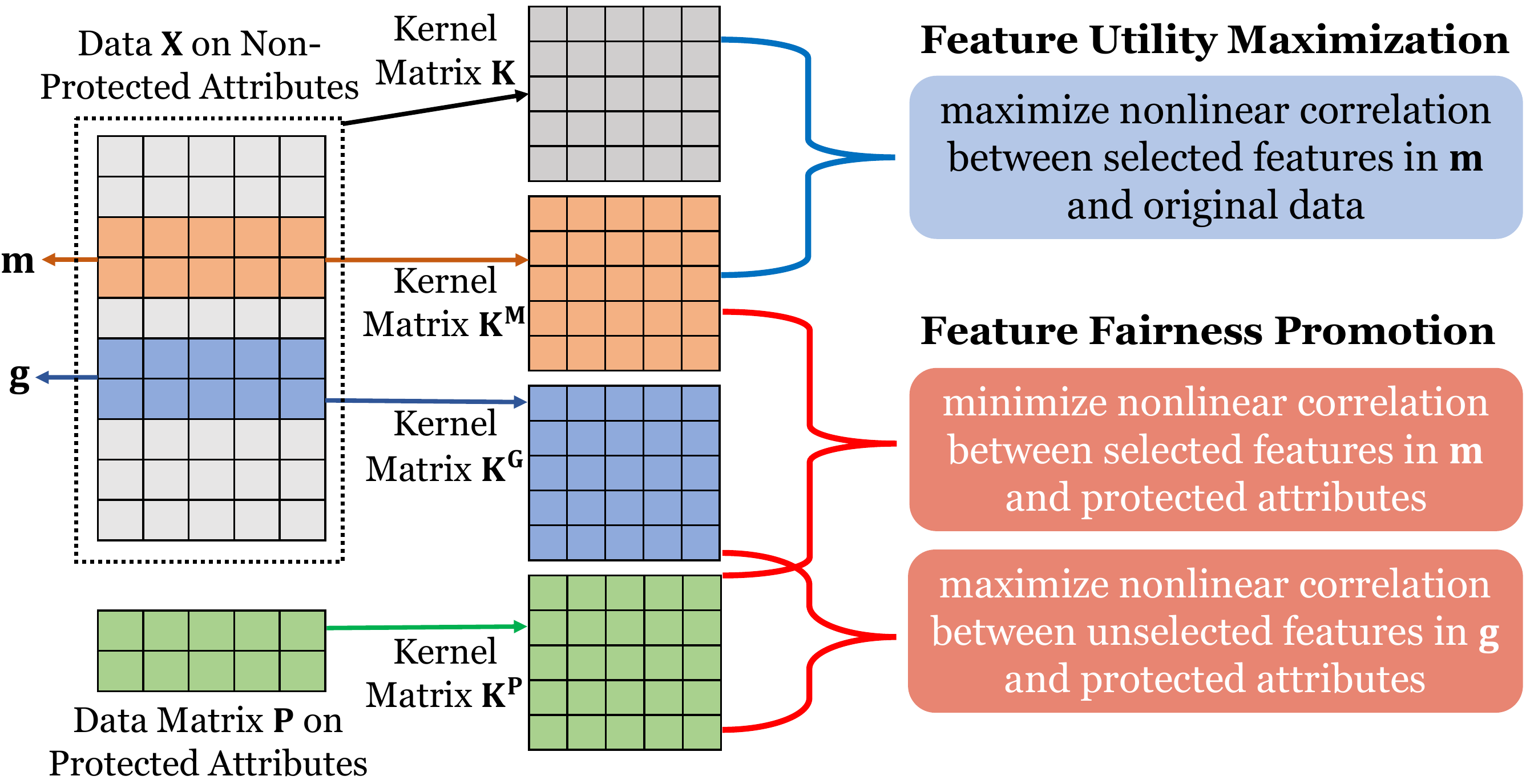}
}
\caption{An illustration of the proposed fairness-aware unsupervised feature selection framework FUFS, which has two components: feature utility maximization and feature fairness promotion.}
\label{framework}
\vspace{-0.15in}
\end{figure}

In this paper, we make an initial investigation of the fairness issues of unsupervised feature selection and develop a general model-agnostic debiasing strategy at the input level. Our proposed efforts have the potential to alleviate unwanted biases before applying downstream learning algorithms and are complementary to the mainstream in-processing algorithmic fairness research~\cite{mehrabi2019survey}. However, developing a fairness-aware unsupervised feature selection framework remains a daunting task, mainly because of the following challenges. Firstly, feature selection should achieve a good trade-off between fairness and feature utility---finding a subset of features that do not exhibit discrimination while still benefiting downstream tasks. However, without label information as supervision signals, we are in short of effective evaluation criteria to quantify these two targets simultaneously. Secondly, due to the trade-off between utility and fairness, it is difficult to simultaneously achieve the maximums of both. Thus, it is necessary to explicitly exclude the fairness-related features from the selected set, which have strong correlations with protected attributes.


To tackle the aforementioned challenges, we propose a novel \underline{F}airness-aware \underline{U}nsupervised \underline{F}eature \underline{S}election (FUFS) framework (as shown in Fig.~\ref{framework}). In essence, to ensure that the selected features do not cause much utility loss for downstream learning algorithms, we select features that can maximally preserve the information in the original feature space. Additionally, we impose fairness constraints to enforce the protected attributes being minimally correlated with the selected features while over associating with a small number of unselected features. All the above considerations are modeled in a joint optimization framework. The major contributions of our work can be summarized as follows:
\begin{itemize}
\setlength\itemsep{0em}
\item We address a crucial and newly emerging problem, fairness-aware unsupervised feature selection, which is essential to debiasing the input data before downstream learning algorithms are involved.

\item We propose a novel FUFS framework, which selects high-quality features by preserving information embedded in the original feature space and obeying the fairness considerations to eliminate sensitive information.

\item We formulate two desiderata of fairness-aware unsupervised feature selection (i.e., feature utility maximization and feature fairness promotion) as an optimization problem with a principled solution.

\item We validate the selected features by feature utility and fairness measurements, where empirical evaluations corroborate the superiority of our proposed framework.
\end{itemize}

\section{Problem Statement}
In this section, we first summarize the notations used in this paper, and then formally define our research problem of \emph{fairness-aware unsupervised feature selection}.

We use bold uppercase letters for matrices (e.g., $\mathbf{A}$), bold lowercase letters for vectors (e.g., $\mathbf{a}$), normal lowercase letters for scalars (e.g., $a$). Also, we use $a_{i}$ to denote the $i$-th element of the vector $\mathbf{a}$, $\mathbf{A}_{*j}$ to denote the $j$-th column of matrix $\mathbf{A}$, $\mathbf{A}_{ij}$ to denote the $i$-th row and $j$-th column of matrix $\mathbf{A}$, and $\text{diag}(\mathbf{a})$ to represent the diagonalization of vector $\mathbf{a}$. Meanwhile, $\mathbf{1}$ denotes a column vector whose elements are all $1$, $\mathbf{I}$ denotes an identity matrix. $\|\mathbf{a}\|_1$ denotes the $\ell_1$-norm of the vector $\mathbf{a}$, respectively. $\|\mathbf{A}\|_{F}$ denotes the Frobenius norm of the matrix $\mathbf{A}$. $\text{Tr}(\mathbf{A})$ denotes the trace of the matrix $\mathbf{A}$ when it is a square matrix.

In this work, we assume there are $n$ data instances, the matrix $\mathbf{P}\in\mathbb{R}^{p\times n}$ denotes the set of $p$ protected attributes for instances (e.g., age, gender, and race), and the matrix $\mathbf{X}\in\mathbb{R}^{d\times n}$ denotes the set of $d$ non-protected attributes (in most cases we have $p\ll d$). The main symbols are summarized in Table~\ref{table:symbols}. With these notations, we formally define our research problem as follows.


\begin{table}[!t]
\small
\caption{Notations and descriptions.}
\vspace{-2mm}
\centering
\begin{tabular}{c|l} \hline
Notation &  Description \\ \hline 
$n$ & Number of instances \\
$d$ & Number of non-protected attributes \\ 
$p$ & Number of protected attributes \\ 
$\mathbf{X}\in\mathbb{R}^{d\times n}$ & Data matrix on non-protected attributes \\ 
$\mathbf{P}\in\mathbb{R}^{p\times n}$ & Data matrix on protected attributes \\ 
$\mathbf{m}\in\{0,1\}^{d}$ & Indicator vector for selected features \\
\multirow{2}{*}{$\mathbf{g}\in\{0,1\}^{d}$} & Indicator vector for features that are highly \\ 
& correlated with protected attributes\\
$\mathbf{M}\in\mathbb{R}^{d\times n}$ & Data subset on the indicator vector $\mathbf{m}$ \\
$\mathbf{G}\in\mathbb{R}^{d\times n}$ & Data subset on the indicator vector $\mathbf{g}$ \\ 
$\mathbf{K}\in\mathbb{R}^{n\times n}$ & Kernel matrix on the input data $\mathbf{X}$\\ 
$\mathbf{K}^{\mathbf{P}}\in\mathbb{R}^{n\times n}$ & Kernel matrix on matrix $\mathbf{P}$ \\ 
$\mathbf{K}^{\mathbf{M}}\in\mathbb{R}^{n\times n}$ & Kernel matrix on matrix $\mathbf{M}$ \\ 
$\mathbf{K}^{\mathbf{G}}\in\mathbb{R}^{n\times n}$ & Kernel matrix on matrix $\mathbf{G}$ \\ \hline
\end{tabular}
\label{table:symbols}
\end{table}

\vspace{0.05in}
\noindent\textbf{Problem Definition} (Fairness-Aware Unsupervised Feature Selection). \textit{Given the input data $\mathbf{X}\in\mathbb{R}^{d\times n}$ and $\mathbf{P}\in\mathbb{R}^{p\times n}$ with $d$ non-protected attributes and $p$ protected attributes, respectively, the problem of fairness-aware unsupervised feature selection aims to select a subset of $k$ features among $d$ non-protected attributes ($k\ll d$) which can maximally preserve the information in the original feature space while being minimally correlated with the protected attributes. } 


\section{The Proposed Framework - FUFS}
In this section, we present our proposed Fairness-aware Unsupervised Feature Selection (FUFS) framework in detail. An overview of the proposed framework is illustrated in Fig. \ref{framework}.

\subsection{Maximizing Feature Utility}
As label information is not available in an unsupervised scenario, we need to seek alternative evaluation criteria to assess the importance of features. One principled metric is capable of ensuring that the selected features can well capture the information embedded in the original feature space. In other words, we would like to maximize the correlation between the selected features and the original ones. However, since the original features could be high-dimensional, complex nonlinear correlations could exist between these two features spaces. Hence, we aim to measure their nonlinear correlation with kernel alignment~\cite{cristianini2006kernel,wei2016nonlinear} techniques. 

Suppose the vector $\mathbf{m}\in\{0,1\}^d$ is the feature selection indicator vector such that $\mathbf{1}^\top\mathbf{m}=k$, where $m_{i}=1$ if the $i$-th feature is selected, otherwise $m_{i}=0$. The data matrix on the selected features can be obtained as $\mathbf{M}=\text{diag}(\mathbf{m})\mathbf{X}$. Then we define a kernel $\kappa$ which implicitly computes the similarity between instances in a high-dimensional reproducing kernel Hilbert space (RKHS)~\cite{aronszajn1950theory}, such that $\mathbf{K}_{ij}=\kappa(\mathbf{X}_{*i},\mathbf{X}_{*j})$ and $\mathbf{K}^{\mathbf{M}}_{ij}=\kappa(\mathbf{M}_{*i},\mathbf{M}_{*j})$. In practice, we can choose polynomial kernel or RBF kernel. Denoting the centering matrix as $\mathbf{H}=\mathbf{I}-\frac{1}{n}\mathbf{1}\mathbf{1}^\top$, these two kernel matrices after centering can be denoted as $\mathbf{K}_{c}=\mathbf{H}\mathbf{K}\mathbf{H}$ and $\mathbf{K}_{c}^{\mathbf{M}}=\mathbf{H}\mathbf{K}_{c}^{\mathbf{M}}\mathbf{H}$, respectively.

With these two centered kernel matrices, we can characterize the inherent nonlinear correlation between these two feature spaces with the centered kernel alignment:
\begin{equation}
\rho\left(\mathbf{K},\mathbf{K}^{\mathbf{M}}\right)=\text{Tr}(\mathbf{K}_{c}\mathbf{K}^{\mathbf{M}}_{c})=\text{Tr}(\mathbf{H}\mathbf{K}\mathbf{H}\mathbf{H}\mathbf{K}^{\mathbf{M}}\mathbf{H}).
\label{eq:utility}
\end{equation}
With the observation that $\mathbf{HH}=\mathbf{H}$ and $\text{Tr}(\mathbf{AB})=\text{Tr}(\mathbf{BA})$ (where $\mathbf{A},\mathbf{B}\in\mathbb{R}^{n\times n}$), we can further simplify $\rho\left(\mathbf{K},\mathbf{K}^{\mathbf{M}}\right)$ as $\text{Tr}(\mathbf{H}\mathbf{K}\mathbf{H}\mathbf{K}^{\mathbf{M}})$. Our goal expects that the selected features in $\mathbf{m}$ can maximally preserve the information embedded in the original feature space.

\subsection{Promoting Feature Fairness}
Although maximizing the centered kernel alignment function in Eq.~(\ref{eq:utility}) helps to select important features that preserve the information of original features, it does not have any fairness considerations such that the selected features may have a risk being associated with the protected attributes in the matrix $\mathbf{P}$. To tackle this issue, we further impose fairness constraints to make the selected features in $\mathbf{M}$ not well aligned with the protected attributes $\mathbf{P}$. To achieve this goal, suppose $\mathbf{K}^{\mathbf{P}}\in\mathbb{R}^{n\times n}$ is the kernel matrix computed from $\mathbf{P}$, we can also leverage centered kernel alignment to minimize the nonlinear correlation between $\mathbf{M}$ and $\mathbf{P}$ in the kernel space: 
\begin{equation}
\rho\left(\mathbf{K}^{\mathbf{M}},\mathbf{K}^{\mathbf{P}}\right)=\text{Tr}(\mathbf{H}\mathbf{K}^{\mathbf{M}}\mathbf{H}\mathbf{K}^{\mathbf{P}}).
\label{eq:fairness1}
\end{equation}
In this way, we can guarantee that the selected features in $\mathbf{M}$ do not have high correlation with the sensitive information.  

Additionally, to further enforcing that the sensitive information is eliminated in the selected features, a small number of unselected features need to exhibit a high degree of correlation with the protected attributes. Toward this goal, we further define a decomposition indicator $\mathbf{g}\in\{0,1\}^{d}$ to indicate the index of non-protected attributes that are highly correlated with $\mathbf{P}$, where $\mathbf{1}^\top\mathbf{g}=l$, and $l$ denotes the number of sensitive features we need to eliminate. Ideally, the nonzero indices of $\mathbf{g}$ should not overlap with those of $\mathbf{m}$. Hence, the data matrix $\mathbf{G}$ corresponding to $\mathbf{g}$ can be obtained as follows:\begin{equation}
\mathbf{G}=\text{diag}(\mathbf{g})(\mathbf{I}-\text{diag}(\mathbf{m}))\mathbf{X}.
\end{equation}
Assume the corresponding kernel matrix is $\mathbf{K}^{\mathbf{G}}\in\mathbb{R}^{n\times n}$, then the centered kernel alignment can also be leveraged to maximize the nonlinear correlation between $\mathbf{G}$ and $\mathbf{P}$:
\begin{equation}
\rho\left(\mathbf{K}^{\mathbf{G}},\mathbf{K}^{\mathbf{P}}\right)=\text{Tr}(\mathbf{H}\mathbf{K}^{\mathbf{G}}\mathbf{H}\mathbf{K}^{\mathbf{P}}).
\label{eq:fairness1}
\end{equation}

\subsection{Objective Function of FUFS}
The above two subsections discuss two desiderata of fairness-aware unsupervised feature selection---preserving the information of original features and imposing fairness constraints. Combining them together, we obtain a joint constrained optimization problem as follows:
\begin{align}
\label{eq:fufs}
\min_{\mathbf{m},\mathbf{g}} &-\text{Tr}\left(\mathbf{HKHK^M}\right)+\alpha\,\text{Tr}\left(\mathbf{HK^MHK^P}\right)\nonumber\\ 
           &-\alpha\,\text{Tr}\left(\mathbf{HK^GHK^P}\right)\\ 
           \text{s.t.}&\;\; \mathbf{m},\mathbf{g}\in\{0,1\}^d, \;\mathbf{1}^\top\mathbf{m}=k,\;\mathbf{1}^\top\mathbf{g}=l. \nonumber
\end{align}
In the above formulation, $\alpha$ is a hyperparameter that can control how strong we would like to enforce the fairness of unsupervised feature selection. 

The optimization problem in Eq.~(\ref{eq:fufs}) is not easy to solve because it is not joint convex w.r.t. $\mathbf{m}$ and $\mathbf{g}$ simultaneously. Although we can employ alternating optimization scheme for a local optimum, the whole optimization still remains difficult due to the discrete nature of variables $\mathbf{m}$ and $\mathbf{g}$. To address this issue, we relax the discrete constraints by reformulating it as a real-valued vector in the range of $[0,1]$. By rewriting the summation constraints $\mathbf{1}^\top\mathbf{m}=k$ and $\mathbf{1}^\top\mathbf{g}=l$ in the form of Lagrangian, we have the following optimization problem: 
\begin{align}
\label{eq:fufs}
\min_{\mathbf{m},\mathbf{g}}&\,\,\mathcal{L}= -\text{Tr}\left(\mathbf{HKHK^M}\right)+\alpha\,\text{Tr}\left(\mathbf{HK^MHK^P}\right)\nonumber\\ 
           &-\alpha\,\text{Tr}\left(\mathbf{HK^GHK^P}\right)+\beta(\|\mathbf{m}\|_1+\|\mathbf{g}\|_1)\\ 
           \text{s.t.}&\;\; \mathbf{m},\mathbf{g}\in[0,1]^d, \nonumber
\end{align}
where the $\ell_1$-norm is introduced for the sparsity of model parameters $\mathbf{m}$ and $\mathbf{g}$. The hyperparameter $\beta$ is used to control the number of selected features that are relevant and do not correlate with protected attributes and the number of unselected features that are highly correlated with protected attributes, respectively.

\noindent \textbf{Updating $\mathbf{m}$ and $\mathbf{g}$.}
We update two model parameters $\mathbf{m}$ and $\mathbf{g}$ alternatively until the objective function converges to a local optimum. The update rules are as follows:
\begin{equation}
    m_{i}\leftarrow P[m_{i}-\eta\,\partial{\mathcal{L}}/\partial m_{i}], \,\, g_{i}\leftarrow P[g_{i}-\eta\,\partial{\mathcal{L}}/\partial g_{i}],
\end{equation}
where $P[x]$ is a box projection operator which projects $x$ into a bounded range. Specifically, since we relax the constraints of $m_i$ and $g_i$ in the range of $[0,1]$, we have $P[x]=0$ if $x<0$, $P[x]=1$ if $x>1$, and otherwise $P[x]=x$. In the above update rules, $\eta$ is the learning rate.

With these updating rules, the pseudo code of the proposed FUFS framework is illustrated in Algorithm~\ref{algorithm:fufs}.

\begin{algorithm}[tb]
\caption{The Proposed Framework FUFS.}
\label{alg:algorithm}
\textbf{Input}: protected attribute matrix $\mathbf{P}\in\mathbb{R}^{p\times n}$, non-protected attribute matrix $\mathbf{X}\in\mathbb{R}^{d\times n}$, number of features $k$ to select, hyperparameters $\alpha, \beta$.\\
\textbf{Output}: the top-$k$ most important features that both preserve the information of original data and do not correlate with protected attributes.
\begin{algorithmic}[1] 
\STATE Initialize indicator vectors $\mathbf{m},\mathbf{g}$ 
\STATE Construct kernel matrices $\mathbf{K}$ from $\mathbf{X}$, $\mathbf{K^P}$ from $\mathbf{P}$, $\mathbf{K^M}$ from $\text{diag}(\mathbf{m})\mathbf{X}$, $\mathbf{K^G}$ from $\text{diag}(\mathbf{g})(\mathbf{I}-\text{diag}(\mathbf{m}))\mathbf{X}$
\WHILE {not convergence}
\STATE  $g_i\leftarrow g_i -\eta\,\partial{\mathcal{L}}/\partial g_i$, $i=1,...,d$
\STATE $g_{i} = \text{min}(1,\text{max}(0,g_{i}))$, $i=1,...,d$
\STATE  $m_i\leftarrow m_i -\eta\,\partial{\mathcal{L}}/\partial m_i$, $i=1,...,d$
\STATE $m_{i} = \text{min}(1,\text{max}(0,m_{i}))$, $i=1,...,d$
\STATE Update matrix $\mathbf{M}$ and kernel matrix $\mathbf{K^M}$
\STATE Update matrix $\mathbf{G}$ and kernel matrix $\mathbf{K^G}$
\ENDWHILE
\STATE Rank features according to the entries in $\mathbf{m}$
\vspace{-0.0in}
\end{algorithmic}
\label{algorithm:fufs}
\end{algorithm}

\section{Experimental Evaluations}
In this section, we conduct experiments on real-world datasets to evaluate the performance of the proposed fairness-aware unsupervised feature selection framework FUFS in terms of both utility and fairness measurements. Before presenting the detailed experiments and findings, we first introduce the experimental settings.

\subsection{Experimental Setup}
\textbf{Datasets}. We perform experiments on four public available datasets. (1) \textsc{Crime}\footnote{http://archive.ics.uci.edu/ml/datasets/Communities+and+\\Crime+Unnormalized}: This dataset combines census data, law enforcement data, and crime data of US communities. We define the percentage of population for African American as a protected attribute. We define two clusters by the number of violent crimes, and the cutoff threshold is 0.15 crimes per 100K population. In total, we have 2,215 communities described by 147 different attributes. (2) \textsc{Adolescent}\footnote{https://www.thearda.com/}: This dataset comes from a longitudinal study of adolescents in Grades 7-12. The attributes are obtained from personal information of the interviewees and their answers to an exhaustive questionnaire. Bio-sex of the interviewee is regarded as the protected attribute and we categorize the interviewees into two clusters by whether their Picture Vocabulary test score is more than 65. In total, this dataset contains 6,504 instances and 2,793 attributes. (3) \textsc{Google+}\footnote{http://snap.stanford.edu/data/ego-Gplus.html}: This dataset comes from Google+, which contains user features and social relations within multiple social circles. Each instance refers to a user and attributes are obtained from personal information of users. Gender is regarded as the protected attribute. We have two clusters defined by the social circles that the users belong to without overlapping. The dataset consists of 2,437 users and 1,695 features. (4) \textsc{Toxicity}\footnote{https://www.kaggle.com/c/jigsaw-unintended-bias-in-toxicity-classification/}: This dataset is obtained from a Toxic Comment Classification Challenge, where each comment is considered as an instance. We apply a tokenizer to transform text data to numerical values. The identity label `female' is regarded as the protected attribute. The features are from identity labels and comment texts. There are two clusters defined by whether the comment is regarded toxic or not. We collect a subset of 200 instances with 4,253 features.



\noindent\textbf{Evaluation Criteria}. For unsupervised feature selection, clustering performance is often used as an evaluation metric~\cite{li2017feature} to assess the quality of selected features. Specifically, we use \emph{Clustering Accuracy (ACC)} and \emph{Normalized Mutual Information (NMI)} to compare the obtained cluster labels with the ground truth cluster labels, and higher values often imply higher quality of selected features. Meanwhile, we use the widely used metrics \emph{Balance}~\cite{li2020deep} and define a new fairness metric \emph{Proportion} as a compliment since \emph{Balance} may be too restrict to reflect the distribution of the clustering. These two metrics are used to quantify how well the selected features can eliminate discrimination---the selected features are considered fairer if they can lead to a more balanced cluster structure toward protected attributes (i.e., higher value of \emph{Balance} and lower value of \emph{Proportion}). These four metrics are defined as follows:
\begin{equation}
\small
ACC=\frac{\sum_{i=1}^{n} \delta \left(y_{i},\textup{map}(\hat{y}_{i})\right)}{n},
\vspace{-0.05in}
\end{equation}
\begin{equation}
\small
NMI=\frac{\sum\limits_{c\in C}\sum\limits_{c'\in C'}p\left(c,c'\right)\log\left(p\left(c,c'\right) / p\left(c\right)p\left(c'\right)\right)}{\textup{mean}\left(H(C),H(C')\right)},
\vspace{-0.05in}
\end{equation}

\begin{equation}
\small
    Balance=\min_{i}\frac{\min_{g}|C_{i}\cap X_{g}|}{|C_{i}|},
\end{equation}

\begin{equation}
\small
    Proportion=\sum_{i}\frac{\max_{g}|C_{i}\cap X_{g}|}{|C_{i}|},
\end{equation}
where $\hat{y}_{i}$ is the clustering result, $y_{i}$ is the true cluster label, $\textup{map}(\cdot)$ is a permutation mapping function that maps $y_{i}$ to the equivalent label from the ground truth and $\delta$ is the indicator function such that $\delta(x,y)=1$ if $x=y$, and $\delta(x,y)=0$ otherwise. $H$ denotes the entropy for a partition set. $C$ and $C'$ denote the obtained clusters and the ground truth, respectively. $C_{i}$ and $X_{g}$ denote the $i$-th cluster and the $g$-th protected subgroup regarding the sensitive attribute. 

\noindent\textbf{Competitive Methods and Implementation}.
We compare our proposed framework FUFS with the following unsupervised feature selection methods: 
\begin{itemize}
\item \textbf{LapScore}~\cite{he2005laplacian}: Laplacian Score selects important features that best align with the local manifold structure of data.
\item \textbf{MCFS}~\cite{cai2010unsupervised}: MCFS selects features that can best preserve the multi-cluster structure of data. 
\item \textbf{UDFS}~\cite{yang2011norm}: UDFS is a pseudo-label based approach that exploits local discriminative information and $\ell_{2,1}$-norm regularization.
\item \textbf{NDFS}~\cite{li2012unsupervised}: NDFS selects important features by performing joint nonnegative spectral analysis and $\ell_{2,1}$-norm regularization.
\item \textbf{REFS}~\cite{li2017reconstruction}: REFS is a reconstruction based approach that learns the reconstruction function from data automatically for unsupervised feature selection.
\end{itemize}We follow the suggestions of the original papers~\cite{he2005laplacian,cai2010unsupervised,yang2011norm,li2012unsupervised,li2017reconstruction} to specify the hyperparameters for these baseline methods. For our proposed FUFS framework, we set the hyperparameters as $\alpha=1, \beta=0.1$ on \textsc{Crime} and \textsc{Google+} while $\alpha=0.01, \beta=10$ on \textsc{Adolescent} and \textsc{Toxicity}. The original distribution of the protected groups in \textsc{Crime} and \textsc{Google+} is more unbalanced thus a larger value of $\alpha$ is necessary to eliminate discrimination. Whereas \textsc{Adolescent} and \textsc{Toxicity} have more features and a larger value of $\beta$ is necessary for unsupervised feature selection. Besides, we specify the kernel function in FUFS as the RBF kernel. For all the compared methods, we first apply unsupervised feature selection to select the top-$k$ ranked features and employ K-means clustering algorithm on the selected features. The clustering results and the ground truth cluster labels are compared and the values on the aforementioned four evaluation metrics can then be obtained. Since the results of K-means depend on initialization, we repeat K-means 50 times and report the average results. Choosing the optimal number of selected features is still an open problem, thus we follow conventional settings~\cite{li2017feature} to vary the number of selected features as $\{10\%, 15\%, 20\%, 25\%, 30\%, 35\%, 40\%\}$ of the total number of features and report the best results regarding different evaluation metrics.

\begin{table}[t]
\caption{Results on \textsc{Crime} w.r.t. cluster validity and fairness.}
\vspace{-0.1in}
	\resizebox{\linewidth}{!}{
\begin{tabular}{c|cc|cc}
\hline
Method & \emph{ACC} & \emph{NMI} & \emph{Balance} & \emph{Proportion} \\ \hline
LapScore & 0.644 (35) & 0.024 (35) & 0.192 (10) & 1.492 (35)\\ 
NDFS & 0.627 (20) & 0.021 (30) & 0.201 (10) & 1.527 (15) \\ 
UDFS & 0.728 (30) & \cellcolor{red!25} 0.150 (30) & 0.107 (40) & \cellcolor{blue!25} 1.456 (25) \\ 
REFS & \cellcolor{red!25} 0.774 (15) & 0.082 (15) & \cellcolor{red!25} 0.208 (25) & 1.552 (40) \\ 
MCFS & 0.683 (25) & 0.101 (20) & 0.182 (20) & 1.511 (25) \\ \hline
FUFS (ours) & \cellcolor{blue!25} 0.758 (15) & \cellcolor{blue!25} 0.141 (35) & \cellcolor{blue!25} 0.204 (35) & \cellcolor{red!25} 1.446 (10)\\ \hline
\end{tabular}
}
\label{crime}
\end{table}

\begin{table}[t]
\caption{Results on \textsc{Adoles.} w.r.t. cluster validity and fairness.}
\vspace{-0.1in}
\centering
	\resizebox{\linewidth}{!}{
\begin{tabular}{c|cc|cc}
\hline
Method & \emph{ACC} & \emph{NMI} & \emph{Balance} & \emph{Proportion} \\ \hline
LapScore & 0.555 (10) & 0.006 (10) & 0.379 (10) & \cellcolor{blue!25} 1.163 (10) \\ 
NDFS & 0.554 (15) & 0.006 (15) & 0.380 (35) &  1.184 (35)\\
UDFS & \cellcolor{blue!25} 0.556 (10) & 0.007 (10) & 0.359 (15) &  1.184 (15) \\
REFS & 0.544 (10) & 0.004 (10) & 0.380 (10) & 1.184 (10) \\ 
MCFS & \cellcolor{red!25} 0.562 (10) & \cellcolor{blue!25} 0.010 (10) & \cellcolor{blue!25} 0.380 (15) & 1.184 (15) \\ \hline
FUFS (ours)& 0.553 (35) & \cellcolor{red!25} 0.013 (35) &  \cellcolor{red!25} 0.407 (10) & \cellcolor{red!25} 1.148 (10)\\ \hline
\end{tabular}
}
\label{Adolescent}
\end{table}

\subsection{Performance Evaluation}
We compare FUFS with different baseline methods in terms of feature utility (\emph{ACC} and \emph{NMI}) and fairness metrics (\emph{Balance} and \emph{Propotion}). The experimental results are shown in Tables 2-5. The number in parentheses denotes the percentage of features when the best performance is achieved. Values in red cell indicates the highest result, and blue cell indicates the second highest one. We make the following observations:
\begin{itemize}
\item FUFS significantly outperforms the baseline methods in terms of \emph{Balance} and \emph{Proportion} with the best performance in almost all cases and the second best performance in terms of \emph{Balance} on \textsc{Crime}. Existing unsupervised feature selection methods often do not have the fairness considerations and deliver the unfair results, while our proposed FUFS framework can obtain the most balanced clustering results across different protected subgroups. 

\item FUFS achieves a good balance between feature utility and feature fairness. While achieving a good performance w.r.t. different fairness metrics, the feature utility is also well maintained as the clustering performance on the selected features is not jeopardized. For example, on \textsc{Crime} and \textsc{Toxicity}, FUFS achieves the second best performance in terms of \emph{ACC} and \emph{NMI} while on \textsc{Adolescent} and \textsc{Google+}, FUFS achieves the best \emph{NMI} values and does not have obvious difference w.r.t. \emph{ACC} compared with the best baseline method.

\item The proposed FUFS framework can achieve great performance in terms of fairness with a small number of features. Specifically, on \textsc{Adolescent} and \textsc{Google+}, FUFS achieves the best results in terms of \emph{Balance} and \emph{Proportion} compared with the baseline methods with merely $10\%$ of the total number of features. On \textsc{Toxicity}, FUFS achieves the best results of fairness with $15\%$ of the total number of features.
\end{itemize}

\begin{table}[t]
\caption{Results on \textsc{Google+} w.r.t. cluster validity and fairness.}
\vspace{-0.1in}
	\resizebox{\linewidth}{!}{
\begin{tabular}{c|cc|cc}
\hline
Method & \emph{ACC} & \emph{NMI} & \emph{Balance} & \emph{Proportion} \\ \hline
LapScore & 0.723 (40) & 0.114 (15) & 0.004 (40) & 1.865 (10)\\
NDFS & \cellcolor{blue!25} 0.724 (40) & 0.113 (40) & 0.000 (10) & 1.885 (15)\\
UDFS & 0.723 (30) & \cellcolor{blue!25} 0.115 (20) & 0.000 (15) & 1.881 (10)\\
REFS & \cellcolor{red!25} 0.724 (35) & 0.114 (20) & 0.004 (20) & 1.886 (15) \\ 
MCFS & 0.719 (40) & 0.109 (15) & \cellcolor{blue!25} 0.228 (10) & \cellcolor{blue!25} 1.412 (35)\\ \hline
FUFS (ours) & 0.721 (10) &\cellcolor{red!25} 0.164 (15) & \cellcolor{red!25} 0.301 (10) & \cellcolor{red!25} 1.308 (10)\\ \hline
\end{tabular}
}
\label{Google}
\end{table}

\begin{table}[t]
\caption{Results on \textsc{Toxicity} w.r.t. cluster validity and fairness.}
\vspace{-0.1in}
	\resizebox{\linewidth}{!}{
\begin{tabular}{c|cc|cc}
\hline
Method & \emph{ACC} & \emph{NMI} & \emph{Balance} & \emph{Proportion} \\ \hline
LapScore & \cellcolor{red!25} 0.803 (30) & \cellcolor{red!25} 0.012 (30) & 0.009 (40) & 1.568 (10)\\ 
NDFS & 0.675 (40) & 0.007 (40) & 0.240 (20) & 1.327 (15) \\
UDFS & 0.663 (40) & 0.006 (30) & 0.284 (10) & \cellcolor{blue!25} 1.309 (10)\\
REFS & 0.674 (40) & 0.007 (35) & \cellcolor{blue!25} 0.334 (40) &  1.579 (40)\\
MCFS & 0.650 (35) & 0.006 (35) & 0.285 (10) & 1.391 (15)\\ \hline
FUFS (ours) & \cellcolor{blue!25} 0.701 (40) & \cellcolor{blue!25} 0.008 (25) & \cellcolor{red!25} 0.409 (15) & \cellcolor{red!25} 1.136 (15) \\ \hline
\end{tabular}
}
\label{Toxicity}
\end{table}

\subsection{In-Depth Exploration of FUFS}
\begin{figure}[!t]
\centering
\resizebox{\linewidth}{!}{
\subfigure[Feature utility]{
\includegraphics[width=4.5cm]{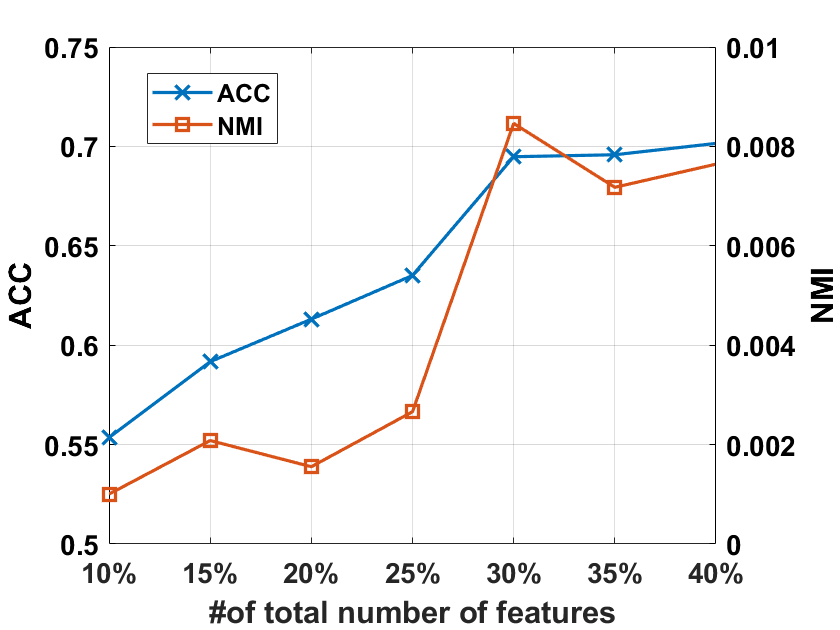}
}
\subfigure[Feature Fairness]{
\includegraphics[width=4.5cm]{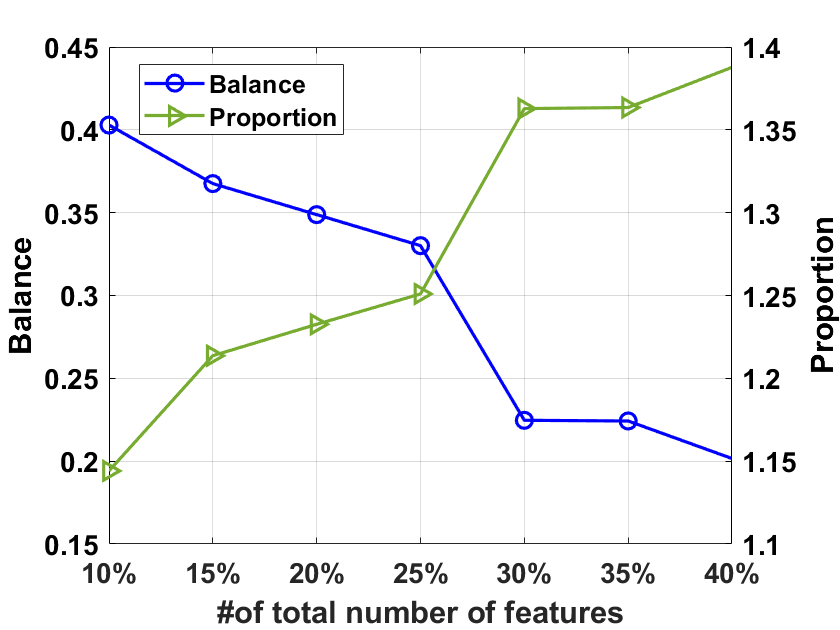}
}
}
\vspace{-0.1in}
\caption{Utility and fairness performance variation on \textsc{Toxicity} w.r.t. different numbers of selected features.}
\label{feanum}\vspace{-4mm}
\end{figure}

\textbf{Effects of the Number of Selected Features}.
Choosing an optimal number of features is still an open problem in unsupervised feature selection research, thus we vary the number of selected features as $\{10\%,15\%,20\%,25\%,30\%,35\%,40\%\}$ of the total feature number and investigate how the feature utility and feature fairness performance change. We only show the results on \textsc{Toxicity} (Fig.~\ref{feanum}) as we have similar observations on other datasets. As we can see, the clustering results (w.r.t. \emph{ACC} and \emph{NMI}) first increase and then keep stable when the number of selected features increase. Meanwhile, the fairness performance (w.r.t. \emph{Balance} and \emph{Proportion}) is the best when only 10\% of features are selected (it should be noted that lower values of \emph{Proportion} denotes fairer results). The fairness performance gradually decreases when the number of selected features continuously increases, the reason is that more features that are correlated with sensitive features could be included in the selected feature subset.

\begin{figure}[!t]
\centering
\subfigure[\textsc{Google+}]{
\includegraphics[width=0.23\textwidth]{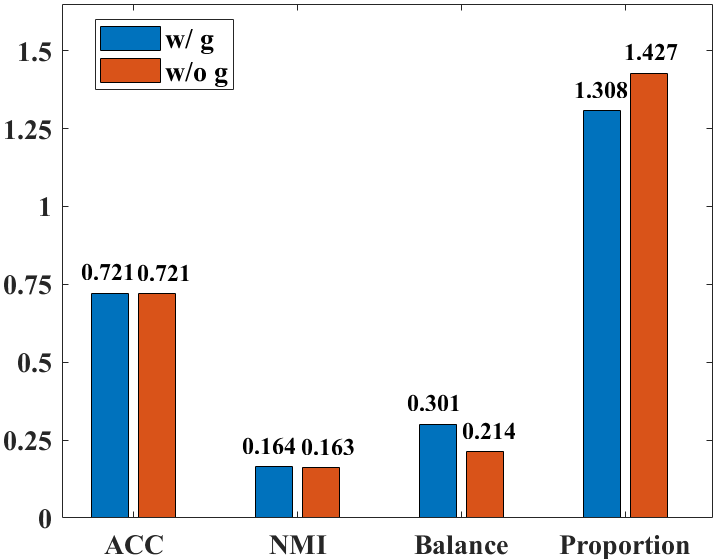}
}\hspace{-2mm}
\subfigure[\textsc{Crime}]{
\includegraphics[width=0.23\textwidth]{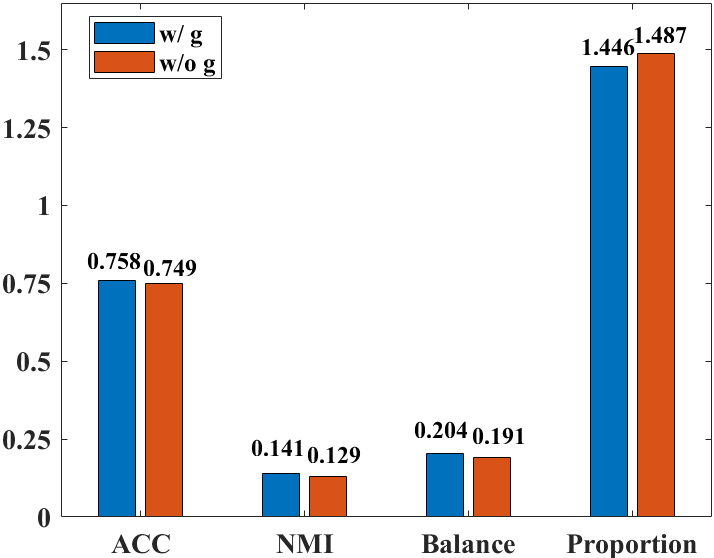}
}\vspace{-4mm}
\caption{Clustering performance w/ and w/o indicator vector $\mathbf{g}$.}\vspace{-4mm}
\label{ablation}
\end{figure}

\noindent\textbf{Effects of the Decomposition Indicator Vector $\mathbf{g}$}.
In order to investigate the effect of the decomposition indicator vector $\textbf{g}$, we remove it from our framework and compare its performance with the original FUFS framework. The results on \textsc{Crime} and \textsc{Google+} shown in Fig.~\ref{ablation} imply that the introduction of the decomposition indicator vector $\textbf{g}$ is necessary and improves both the utility and fairness performance. We also compare the fairness performance based on the top-ranked features in $\textbf{m}$ and $\textbf{g}$ and the results are shown in Table \ref{vector_g}. The number in parentheses denotes the percentage of features when the best performance is achieved. It is obvious that the clustering results based on the top-ranked features in the vector $\textbf{m}$ are more fair than those in the vector $\textbf{g}$. It shows the effectiveness of introducing the decomposition indicator vector $\textbf{g}$, which can help eliminate the sensitive information in the selected feature subset. 


\begin{table}[!t]
\caption{Fairness results (w.r.t. \emph{Balance} and \emph{Proportion}) comparison based on the top-ranked features in \textbf{m} and \textbf{g}.}\vspace{-2mm}
\resizebox{\linewidth}{!}{
\begin{tabular}{c|cc|cc}
\hline
 & \multicolumn{2}{c|}{Top-$k$ ranked features in \textbf{m}} & \multicolumn{2}{c}{Top-$k$ ranked features in \textbf{g}} \\ \hline
Dataset & \emph{Balance} & \emph{Proportion} & \emph{Balance} & \emph{Proportion} \\ \hline
\textsc{Crime} & 0.204 (35) & 1.446 (10) & 0.188 (40) & 1.468 (15) \\
\textsc{Adolescent} & 0.407 (10) & 1.148 (10) & 0.308 (40) & 1.179 (30) \\
\textsc{Google+} & 0.301 (10) & 1.308 (10) & 0.000 (20) & 1.863 (20) \\
\textsc{Toxicity} & 0.409 (15) & 1.136 (15) & 0.063 (15) & 1.629 (40)\\ \hline
\end{tabular}}\label{vector_g}
\end{table}

\begin{figure}[]
    \centering
    \resizebox{\linewidth}{!}{
    \subfigure[\textit{ACC}]{
    \includegraphics[width=4.5cm]{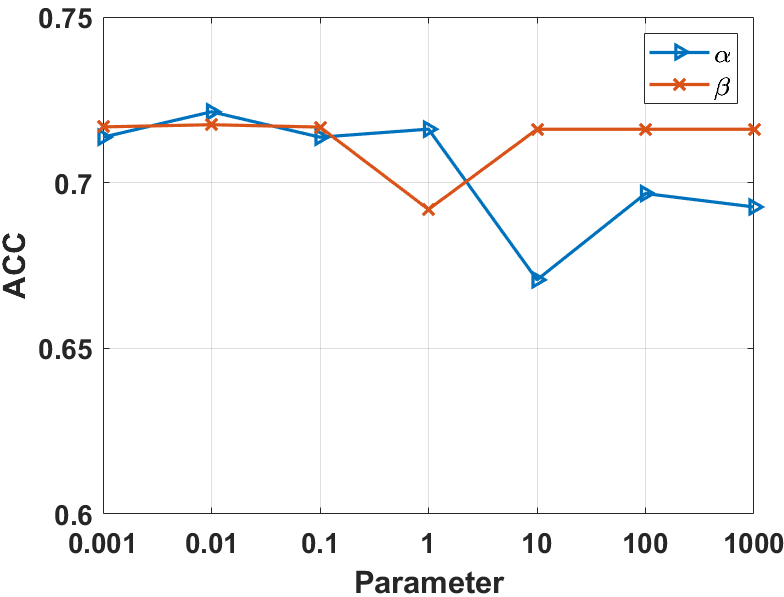}
    }
    \subfigure[\textit{Balance}]{
    \includegraphics[width=4.5cm]{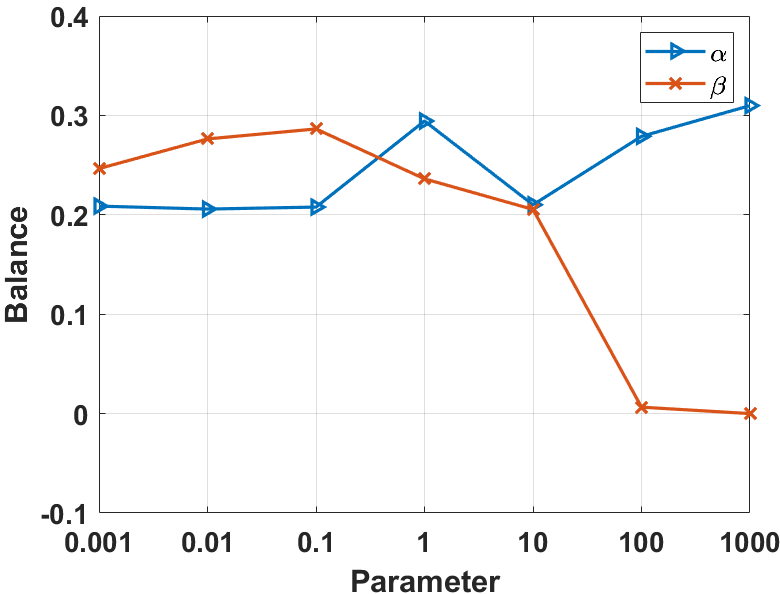}
    }
    }\vspace{-4mm}
    \caption{Performance variation on \textsc{Google+} w.r.t. different parameter settings. X-axis is not in a linear scale.}
    \label{fig:para}\vspace{-4mm}
\end{figure}

\noindent\textbf{Parameter Study}. The proposed framework FUFS has two important hyperparameters. The first parameter $\alpha$ controls how strong we would like to enforce the fairness of unsupervised feature selection. The other parameter $\beta$ controls the sparsity of the proposed model. We first fix $\beta=0.1$ and then vary $\alpha$ among $\{0.001,0.01,0.1,1,10,100,1000\}$. Next, We first fix $\alpha=1$ and then vary $\beta$ among $\{0.001,0.01,0.1,1,10,100,1000\}$. The performance on \textsc{Google+} is shown in Fig.~\ref{fig:para}. It should be noted that as the X-axis is plotted in a log scale, we do not expect to see a smooth curve. Due to space limit, we only show the parameter study results on \textsc{Google+} in terms of \emph{ACC} and \emph{Balance}. The results imply that the clustering performance is relatively stable when $\alpha=1, \beta \in [0.001, 0.1]$ or $\alpha\in[0.001, 0.1], \beta=0.1$. When the parameter $\alpha$ increases, the algorithm becomes more partial to the fairness consideration with decreasing \emph{ACC} and increasing \emph{Balance}. It can be observed that the fairness performance decreases a lot if $\beta$ is specified as a very large value.

\vspace{-0.05in}
\section{Related Work}

\noindent \textbf{Unsupervised Feature Selection.}
Due to the expensive annotation cost, unsupervised feature selection has sparked great interests in recent years. To quantify the importance of features, unsupervised methods often rely on alternative evaluation criteria based on different characteristics of data. Specifically, similarity based methods~\cite{he2005laplacian,zhao2007spectral} select features that can best preserve the local manifold structure of data. Another family of methods aim to select important features that can best reconstruct~\cite{Farahat,li2017reconstruction} or maximally preserve information embedded in the original features~\cite{wei2016nonlinear}. A number of studies learn the pseudo label from data by exploiting local/global discriminative information and select features to predict these pseudo labels with $\ell_{2,1}$-norm based regression~\cite{li2012unsupervised,li2018unsupervised}.  Recently, data reconstruction~\cite{li2017reconstruction,Farahat} emerged as a new criterion to evaluate feature relevance, which evaluates the capability of features in approximating the original data through a reconstruction function.

\noindent \textbf{Fairness of Unsupervised Learning Methods.} To our best knowledge, we are the first to study the fairness issue of unsupervised feature selection. Here we review some related fairness topics in terms of clustering and representation learning. The initial work~\cite{chierichetti2017fair} defines fair variants of classical clustering problems such as $k$-center and $k$-median and proposes the concepts of fairlets and fairlet decomposition, which is further extended to $k$-means++ algorithm by~\cite{schmidt2018fair}. Other related works focus on scalable fair clustering~\cite{backurs2019scalable}, fair spectral clustering~\cite{kleindessner2019guarantees}, and deep fair clustering~\cite{li2020deep}. Another family of work aims to learn fair representations. Fair PCA~\cite{creager2019flexibly} is a two-step algorithm for dimension reduction. Fair Autoencoders~\cite{louizos2015variational,moyer2018invariant} encourage independence between sensitive and latent factors of variation for representation learning. Extended work~\cite{creager2019flexibly} learns general-purpose flexible fair representations regarding multiple sensitive attributes.

\vspace{-0.05in}
\section{Conclusion}
Unsupervised feature selection plays an essential role in preparing high-dimensional and unlabeled data for various learning tasks and has been increasingly used in high-stake applications. Despite its fundamental importance, the fairness of unsupervised feature selection has largely remained nascent. In this paper, we addressed a novel problem of fairness-aware unsupervised feature selection and developed a principled framework FUFS. The proposed framework leverages the technique of kernel alignment to select high-quality features that achieve a good balance between improving downstream learning tasks and eliminating sensitive information that is highly correlated with protected attributes. These two desiderata were modeled together in a joint optimization framework. Experimental evaluations on real-world datasets demonstrated the superiority of the proposed FUFS framework in terms of feature utility and feature fairness. 





\newpage
\bibliographystyle{named}
\bibliography{ijcai21}

\end{document}